\documentclass[12pt]{article}
\usepackage{amsmath}
\usepackage{amsfonts}
\usepackage{geometry}
\usepackage{hyperref}
\usepackage{booktabs}
\usepackage{graphicx}
\usepackage{float}
\usepackage{algorithm}
\usepackage{algpseudocode}
\begin{document}

\title{Dimer-Enhanced Optimization: A First-Order Approach to Escaping Saddle Points in Neural Network Training}
\author{
    Yue Hu$^{1}$, Zanxia Cao$^{2}$, Yingchao Liu$^{3}$ \\
    \small $^{1}$School of Bioengineering, Qilu University of Technology (Shandong Academy of Sciences), \\
    \small No. 3501 Daxue Road, Jinan, Shandong, China \\
    \small $^{2}$Shandong Provincial Key Laboratory of Biophysics, Institute of Biophysics, \\
    \small Dezhou University, Dezhou 253023, China \\
    \small $^{3}$Shandong Provincial Hospital, Shandong First Medical University \\
    \small Email: huyue@qlu.edu.cn, 303004955@qq.com, yingchaoliu@email.sdu.edu.cn
}
\date{July 26, 2025}

\maketitle

\begin{abstract}
First-order optimization methods, such as SGD and Adam, are widely used for training large-scale deep neural networks due to their computational efficiency and robust performance. However, relying solely on gradient information, these methods often struggle to navigate complex loss landscapes featuring flat regions, plateaus, and saddle points. Second-order methods, which utilize curvature information from the Hessian matrix, can address these challenges but are computationally infeasible for large models. The Dimer method, a first-order technique that constructs two closely spaced points to probe the local geometry of a potential energy surface, offers an efficient approach to estimate curvature using only gradient information. Drawing inspiration from its application in molecular dynamics simulations for locating saddle points, we propose Dimer-Enhanced Optimization (DEO), a novel framework to facilitate escaping saddle points in neural network training. Unlike its original use, DEO adapts the Dimer method to explore a broader region of the loss landscape, efficiently approximating the Hessian’s smallest eigenvector without computing the full Hessian matrix. By periodically projecting the gradient onto the subspace orthogonal to the minimum curvature direction, DEO guides the optimizer away from saddle points and flat regions, promoting more effective training while significantly reducing sampling time costs through non-stepwise updates. Preliminary experiments on Transformer-based toy models show that DEO achieves competitive performance compared to standard first-order methods, enabling improved navigation of complex loss landscapes. Our work offers a practical approach to repurpose physics-inspired, first-order curvature estimation for enhancing neural network training in high-dimensional spaces.
\end{abstract}

\section{Introduction}

Training deep neural networks involves optimizing high-dimensional, non-convex loss functions, often hindered by complex loss landscapes featuring flat regions, narrow valleys, and saddle points~\cite{dauphin2014saddle}. First-order methods, such as Stochastic Gradient Descent (SGD)~\cite{robbins1951sgd}, Adam~\cite{kingma2014adam}, and AdamW~\cite{loshchilov2017adamw}, are widely adopted for their computational efficiency and robust performance. These methods rely on local gradient information, leading to slow convergence in ill-conditioned regions or stagnation near saddle points where the gradient is near zero but not at a local minimum.

Second-order methods leverage the Hessian matrix’s curvature information via the Newton step \(\Delta \mathbf{\theta} = -\mathbf{H}^{-1} \mathbf{g}\), but their \(\mathcal{O}(N^2)\) memory and \(\mathcal{O}(N^3)\) complexity make them infeasible for large models. Recent methods like Sophia~\cite{liu2023sophia} approximate the Hessian’s diagonal with gradient differences, achieving low overhead but missing non-diagonal interactions critical for saddle points.

Inspired by the Dimer method from molecular dynamics~\cite{henkelman1999dimer}, which probes potential energy surfaces to locate saddle points, we propose Dimer-Enhanced Optimization (DEO). DEO adapts the Dimer method to escape saddle points and stabilize neural network training, using a displacement \(\Delta R \approx 10 \cdot \text{lr}\) to probe broader loss landscapes and periodically correcting gradients of optimizers like Adam. Experiments on Transformer-based models demonstrate DEO’s competitive performance and its ability to mitigate training instabilities, offering a practical approach for more robust optimization.

\section{Background and Related Work}

DEO integrates first-order optimization, curvature estimation, and saddle point escape strategies.

\subsection{First-Order Stochastic Optimizers}

SGD~\cite{robbins1951sgd} updates parameters using mini-batch gradients but converges slowly in complex landscapes. Adam~\cite{kingma2014adam} combines momentum and adaptive learning rates, while AdamW~\cite{loshchilov2017adamw} improves generalization with decoupled weight decay. Sophia~\cite{liu2023sophia} approximates the Hessian’s diagonal with gradient differences, achieving low overhead. These methods (SGD, Adam, AdamW, Sophia) are our baselines.

\subsection{Our Contribution}

DEO enhances first-order optimizers by:
- Using the Dimer method for non-diagonal curvature, unlike gradient-only SGD, Adam, AdamW.
- Targeting the Hessian’s smallest eigenvector, unlike Sophia’s diagonal focus.
- Employing \(\Delta R \approx 10 \cdot \text{lr}\) and periodic updates for efficiency.

\section{Methodology}

We develop Dimer-Enhanced Optimization (DEO) by adapting the Dimer method~\cite{henkelman1999dimer} to escape saddle points, aligning with its molecular dynamics formulation.

\subsection{Preliminaries: The Dimer Method}

The Dimer method locates saddle points in molecular dynamics by probing potential energy surfaces~\cite{henkelman1999dimer}. For a loss function \(L(\mathbf{\theta})\), DEO adapts this to estimate the Hessian’s smallest eigenvector \(\hat{\mathbf{N}}\) using gradients. We construct one additional point with \(\Delta R \approx 10 \cdot \text{lr}\) (e.g., \(\Delta R = 6 \times 10^{-3}\) for \(\text{lr} = 6 \times 10^{-4}\)) to probe a broader loss landscape, using the current point’s gradient.

Given \(\mathbf{\theta}\) and \(\mathbf{g} = \nabla L(\mathbf{\theta})\), we construct:
\[
\mathbf{\theta}_2 = \mathbf{\theta} + \Delta R \hat{\mathbf{N}}
\]
with gradient \(\mathbf{g}_2 = \nabla L(\mathbf{\theta}_2)\). The rotational force updates \(\hat{\mathbf{N}}\):
\[
\mathbf{F}_R = (\mathbf{g}_2 - \mathbf{g}) - \big((\mathbf{g}_2 - \mathbf{g}) \cdot \hat{\mathbf{N}}\big) \hat{\mathbf{N}}
\]
\[
\hat{\mathbf{N}}_{t+1} = \text{normalize}(\hat{\mathbf{N}}_t + \eta_{\text{rot}} \mathbf{F}_R)
\]
where \(\eta_{\text{rot}} = 10^{-3}\). The curvature is:
\[
C = \frac{L(\mathbf{\theta}_2) - L(\mathbf{\theta})}{\Delta R}
\]
A negative \(C\) confirms alignment with the minimum curvature direction, requiring one extra gradient evaluation~\cite{henkelman1999dimer}.

\subsection{Gradient Correction}

DEO corrects the gradient by projecting it orthogonal to \(\hat{\mathbf{N}}\), with coefficient \(\alpha\) (e.g., 5.0):
\[
\mathbf{g}_{\text{mod}} = \mathbf{g} - \alpha (\mathbf{g} \cdot \hat{\mathbf{N}}) \hat{\mathbf{N}}
\]
This removes low-curvature components, guiding the optimizer away from saddle points.

\subsection{Dimer-Enhanced Optimization (DEO) with Adam}

DEO enhances the Adam optimizer~\cite{kingma2014adam} with periodic Dimer corrections. Adam updates parameters using:
\[
\mathbf{m}_t = \beta_1 \mathbf{m}_{t-1} + (1 - \beta_1) \mathbf{g}_t, \quad \mathbf{v}_t = \beta_2 \mathbf{v}_{t-1} + (1 - \beta_2) \mathbf{g}_t^2
\]
\[
\hat{\mathbf{m}}_t = \frac{\mathbf{m}_t}{1 - \beta_1^t}, \quad \hat{\mathbf{v}}_t = \frac{\mathbf{v}_t}{1 - \beta_2^t}
\]
\[
\mathbf{\theta}_{t+1} = \mathbf{\theta}_t - \text{lr} \cdot \frac{\hat{\mathbf{m}}_t}{\sqrt{\hat{\mathbf{v}}_t} + \epsilon}
\]
where \(\beta_1 = 0.9\), \(\beta_2 = 0.95\), \(\epsilon = 10^{-8}\). DEO replaces \(\mathbf{g}_t\) with \(\mathbf{g}_{\text{mod}}\) from the Dimer correction, using \(\Delta R \approx 10 \cdot \text{lr}\). Algorithm~\ref{alg:deo} details the integration.

\begin{algorithm}
\caption{Dimer-Enhanced Optimization (DEO) with Adam}\label{alg:deo}
\begin{algorithmic}[1]
\State \textbf{Input}: Loss function $L(\mathbf{\theta})$, learning rate $\text{lr}$, update frequency $f$, displacement $\Delta R \approx 10 \cdot \text{lr}$, step size $\eta_{\text{rot}}$, correction coefficient $\alpha$, Adam parameters $\beta_1$, $\beta_2$, $\epsilon$
\State Initialize $\mathbf{\theta}$, $\hat{\mathbf{N}}_{\text{cached}} \gets$ random unit vector, $\mathbf{m}_0 \gets \mathbf{0}$, $\mathbf{v}_0 \gets \mathbf{0}$
\For{each step $t$}
    \State Compute gradient $\mathbf{g} \gets \nabla L(\mathbf{\theta})$
    \If{$t \mod f = 0$} \Comment{Expensive step}
        \State $\mathbf{\theta}_2 \gets \mathbf{\theta} + \Delta R \hat{\mathbf{N}}_{\text{cached}}$
        \State $\mathbf{g}_2 \gets \nabla L(\mathbf{\theta}_2)$
        \State $\mathbf{F}_R \gets (\mathbf{g}_2 - \mathbf{g}) - \big((\mathbf{g}_2 - \mathbf{g}) \cdot \hat{\mathbf{N}}_{\text{cached}}\big) \hat{\mathbf{N}}_{\text{cached}}$
        \State $\hat{\mathbf{N}}_{\text{new}} \gets \text{normalize}(\hat{\mathbf{N}}_{\text{cached}} + \eta_{\text{rot}} \mathbf{F}_R)$
        \State $\hat{\mathbf{N}}_{\text{cached}} \gets \hat{\mathbf{N}}_{\text{new}}$
        \State $\mathbf{g}_{\text{mod}} \gets \mathbf{g} - \alpha (\mathbf{g} \cdot \hat{\mathbf{N}}_{\text{new}}) \hat{\mathbf{N}}_{\text{new}}$
    \Else \Comment{Cheap step}
        \State $\mathbf{g}_{\text{mod}} \gets \mathbf{g} - \alpha (\mathbf{g} \cdot \hat{\mathbf{N}}_{\text{cached}}) \hat{\mathbf{N}}_{\text{cached}}$
    \EndIf
    \State $\mathbf{m}_t \gets \beta_1 \mathbf{m}_{t-1} + (1 - \beta_1) \mathbf{g}_{\text{mod}}$
    \State $\mathbf{v}_t \gets \beta_2 \mathbf{v}_{t-1} + (1 - \beta_2) \mathbf{g}_{\text{mod}}^2$
    \State $\hat{\mathbf{m}}_t \gets \mathbf{m}_t / (1 - \beta_1^t)$, $\hat{\mathbf{v}}_t \gets \mathbf{v}_t / (1 - \beta_2^t)$
    \State $\mathbf{\theta}_{t+1} \gets \mathbf{\theta}_t - \text{lr} \cdot \hat{\mathbf{m}}_t / (\sqrt{\hat{\mathbf{v}}_t} + \epsilon)$
\EndFor
\end{algorithmic}
\end{algorithm}

\section{Experiments}

We evaluate DEO using two Transformer-based toy models of varying complexity, inspired by the nanoGPT framework~\cite{nanoGPT}. The experiments are conducted on a large public text corpus and compare DEO-enhanced optimizers against their standard counterparts.

\subsection{Experiment 1: Simpler Language Model}

\subsubsection{Experimental Setup}

\begin{itemize}
    \item \textbf{Model}: A simpler language model with 6 Transformer layers, 4 attention heads, and a hidden dimension of 256. The model is configured for sequence classification.
    \item \textbf{Dataset}: A large public text corpus, with a micro-batch size of 12 and 20 gradient accumulation steps (total batch size of 240).
    \item \textbf{Optimizers}: DEO (with Adam, AdamW, SGD, and Sophia bases), Adam, AdamW, SGD, and Sophia.
    \item \textbf{Hyperparameters}: Learning rate \(\text{lr} = 6 \times 10^{-4}\) with cosine decay, \(\Delta R = 6 \times 10^{-3}\), \(f=10\), and \(\alpha = 5.0\).
\end{itemize}

\subsubsection{Results}

\begin{figure}[H]
    \centering
    \includegraphics[width=\textwidth]{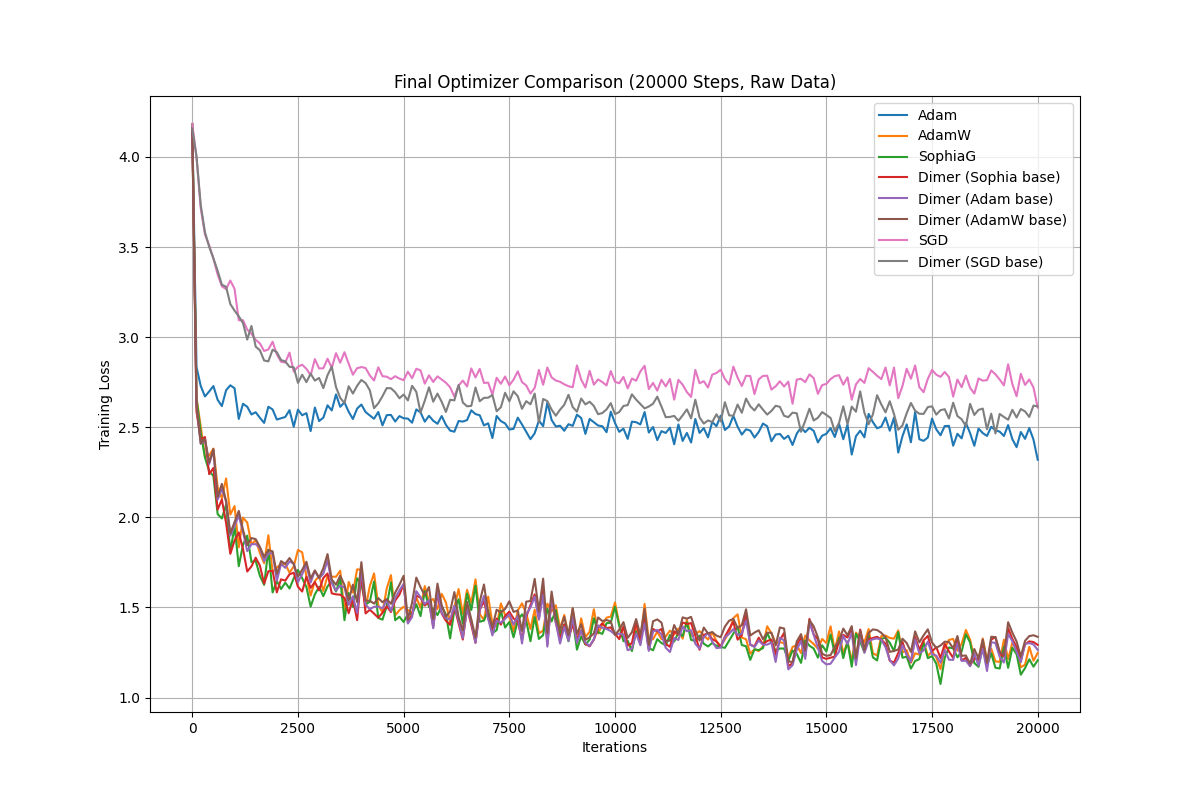}
    \caption{Raw (unsmoothed) training loss comparison on a simpler language toy model. DEO-enhanced variants show competitive performance, particularly with Adam and AdamW bases.}
    \label{fig:optimizer_comparison}
\end{figure}

As shown in Figure~\ref{fig:optimizer_comparison}, the DEO-enhanced optimizers performed competitively. The loss curves for DEO variants closely tracked or slightly outperformed their baselines, particularly the Adam and AdamW bases. While SGD-based optimizers converged more slowly, the Dimer-enhanced version achieved a slightly lower final loss, indicating a consistent, albeit small, benefit from the gradient correction. Sophia also performed strongly, but the results show that DEO provides a comparable level of performance in this setting.

\subsection{Experiment 2: More Complex Language Model}

\subsubsection{Experimental Setup}

\begin{itemize}
    \item \textbf{Model}: A more complex language model with 8 Transformer layers, 8 attention heads, and an embedding dimension of 256, trained from scratch for language modeling.
    \item \textbf{Dataset}: The same public text corpus, using an identical data loading strategy.
    \item \textbf{Optimizers}: DEO-enhanced versions of Adam, AdamW, and SGD were compared against their standard counterparts.
    \item \textbf{Hyperparameters}: The hyperparameters were kept consistent with the first experiment, including a learning rate of \(6 \times 10^{-4}\) with cosine decay, \(\Delta R = 6 \times 10^{-3}\), and \(\alpha = 5.0\).
\end{itemize}

\subsubsection{Results}

\begin{figure}[H]
    \centering
    \includegraphics[width=\textwidth]{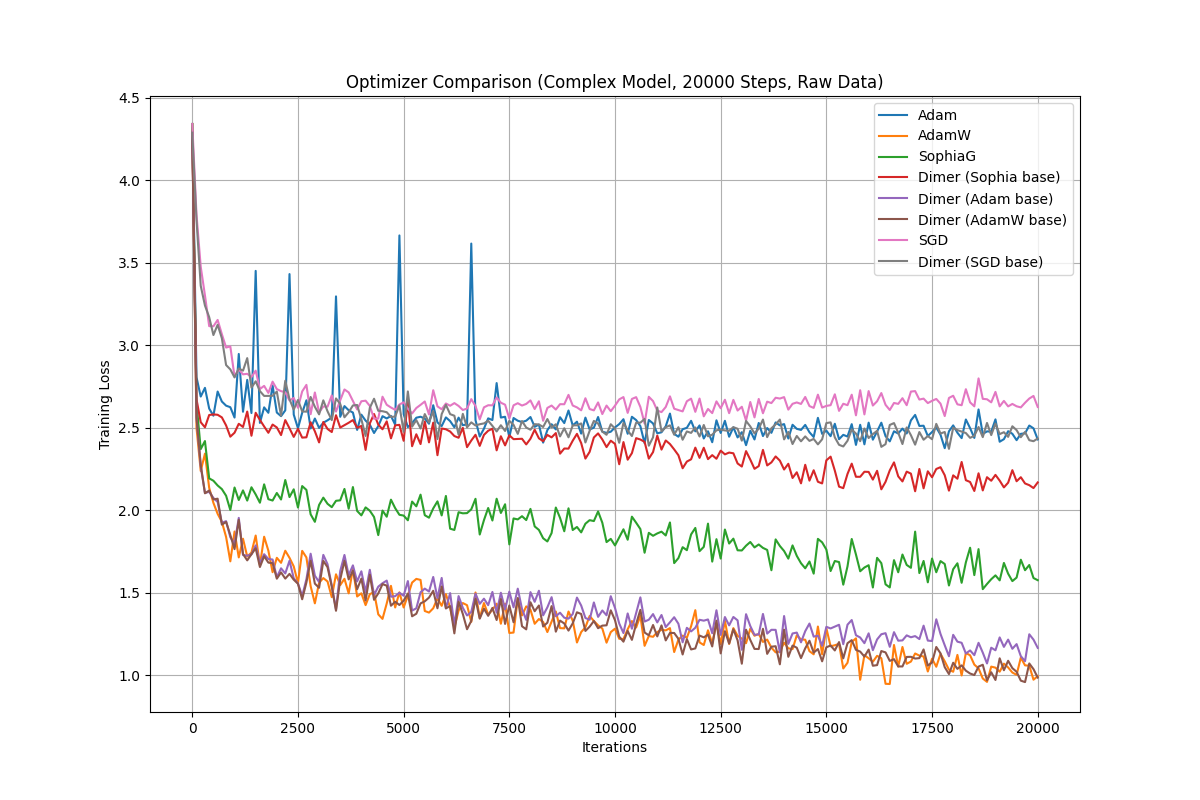}
    \caption{Raw (unsmoothed) training loss comparison on a more complex language model. The Dimer correction notably stabilizes the standard Adam optimizer and improves the performance of AdamW.}
    \label{fig:optimizer_comparison_complex}
\end{figure}

The results for the more complex model are presented in Figure~\ref{fig:optimizer_comparison_complex}. Here, the benefits of DEO become more apparent. The standard Adam optimizer exhibited significant training instability, characterized by sharp loss spikes. The DEO enhancement (Dimer Adam base) completely smoothed out these instabilities, leading to robust and stable convergence. DEO-enhanced AdamW also achieved a lower final loss than its baseline. These results strongly suggest that DEO’s gradient correction is particularly effective in stabilizing adaptive optimizers in more complex and potentially pathological loss landscapes.

\section{Discussion and Future Work}

\subsection{The Role of Non-Diagonal Curvature in Stabilization}

Our experiments highlight a key strength of DEO: its ability to stabilize training in complex models. As shown in Figure~\ref{fig:optimizer_comparison_complex}, the standard Adam optimizer exhibited significant loss spikes, a sign of instability. The DEO-enhanced Adam variant smoothed these instabilities, leading to robust and monotonic convergence. This suggests that the non-diagonal curvature information captured by the Dimer method is crucial for navigating highly variable or pathological regions of the loss landscape. While diagonal methods like Sophia are efficient, they may miss the parameter interactions that DEO’s gradient correction, guided by the Hessian's estimated minimum eigenvector, successfully addresses.

\subsection{Synergy with Adaptive Optimizers}

The results from both experiments consistently show that DEO provides the most significant benefits when paired with adaptive optimizers like Adam and AdamW. For SGD and Sophia, the improvements were marginal. This suggests a synergistic relationship: the Dimer correction guides the optimizer out of saddle points or flat regions, while the adaptive learning rates and momentum of Adam(W) allow it to effectively exploit the new, more promising directions. The gradient projection appears to be most effective when the underlying optimizer can dynamically adjust its step sizes.

\subsection{Limitations}

Despite its promising results, our study has several limitations. First, the hyperparameters for DEO—update frequency \(f\), correction coefficient \(\alpha\), and displacement \(\Delta R\)—require careful tuning. Second, our experiments were confined to two Transformer-based toy models and a single dataset. The performance of optimizers is known to be highly sensitive to the specific model architecture and data distribution. As our results show, the effectiveness of an optimizer can vary significantly between a simpler and a more complex model, suggesting that there is a component of happenstance in achieving state-of-the-art results. Therefore, the generalization of our findings to other types of models and datasets needs further investigation. The computational overhead of the periodic Dimer calculation, while minimal, also remains an added cost.

\subsection{Future Work}

Our findings open several avenues for future research:
\begin{itemize}
    \item \textbf{Adaptive DEO}: The fixed hyperparameters could be made adaptive. For instance, \(\alpha\) could be dynamically adjusted based on the measured curvature or training stability, applying stronger corrections only when needed. The update frequency \(f\) could also be scheduled to decrease as training progresses.
    \item \textbf{Hybrid Optimization Strategies}: A hybrid approach combining DEO with Sophia could be beneficial. DEO could be activated selectively when signs of instability (e.g., loss spikes or high variance in gradients) are detected, while relying on Sophia’s efficient diagonal approximation for stable training phases.
    \item \textbf{Large-Scale Evaluation}: The most critical next step is to evaluate DEO’s performance and scalability on larger, more complex models, such as GPT-3 scale models, to confirm if the benefits observed in our experiments hold true in large-scale settings.
\end{itemize}

\section*{Code Availability}
The source code and experimental data for this paper are publicly available at: \url{https://github.com/YueHuLab/DimerTrainer}.

\section{Conclusion}

Dimer-Enhanced Optimization (DEO) successfully adapts the Dimer method from molecular dynamics to address the challenge of escaping saddle points in neural network training. By using a single-point Dimer construction to estimate the minimum curvature direction and projecting the gradient away from it, DEO effectively enhances standard first-order optimizers with minimal computational overhead. Our experiments on Transformer-based models demonstrate that DEO, particularly when combined with adaptive optimizers like Adam and AdamW, not only achieves competitive convergence but also significantly improves training stability by mitigating loss spikes. This work provides strong evidence that incorporating non-diagonal curvature information is a practical and effective strategy for robust optimization in high-dimensional, non-convex landscapes, offering a valuable bridge between first- and second-order methods.


\begin{thebibliography}{99}
\bibitem{dauphin2014saddle}
Dauphin, Y. N., Pascanu, R., Gulcehre, C., Cho, K., Ganguli, S., \& Bengio, Y. (2014). Identifying and attacking the saddle point problem in high-dimensional non-convex optimization. In \textit{Advances in Neural Information Processing Systems (NeurIPS)} (pp. 2933--2941). \href{https://arxiv.org/abs/1406.2572}{arXiv:1406.2572}

\bibitem{robbins1951sgd}
Robbins, H., \& Monro, S. (1951). A stochastic approximation method. \textit{The Annals of Mathematical Statistics}, 22(3), 400--407. \href{https://doi.org/10.1214/aoms/1177729586}{https://doi.org/10.1214/aoms/1177729586}

\bibitem{kingma2014adam}
Kingma, D. P., \& Ba, J. (2015). Adam: A method for stochastic optimization. In \textit{3rd International Conference on Learning Representations (ICLR)}. \href{https://arxiv.org/abs/1412.6980}{arXiv:1412.6980}

\bibitem{loshchilov2017adamw}
Loshchilov, I., \& Hutter, F. (2019). Decoupled weight decay regularization. In \textit{7th International Conference on Learning Representations (ICLR)}. \href{https://arxiv.org/abs/1711.05101}{arXiv:1711.05101}

\bibitem{liu2023sophia}
Liu, H., Li, Z., Que, D., Zhang, X., \& Feng, J. (2023). Sophia: A scalable stochastic second-order optimizer for language model pre-training. In \textit{International Conference on Learning Representations (ICLR)}. \href{https://arxiv.org/abs/2305.12851}{arXiv:2305.12851}

\bibitem{henkelman1999dimer}
Henkelman, G., \& J\'onsson, H. (1999). A dimer method for finding saddle points on high dimensional potential surfaces using only first derivatives. \textit{The Journal of Chemical Physics}, 111(15), 7010--7022. \href{https://doi.org/10.1063/1.480097}{https://doi.org/10.1063/1.480097}

\bibitem{nanoGPT}
Karpathy, A. (2022). nanoGPT. GitHub repository. \href{https://github.com/karpathy/nanoGPT}{https://github.com/karpathy/nanoGPT}

\end{thebibliography}
\end{document}